\documentclass[11pt]{article}

\usepackage[final]{acl}

\usepackage{times}
\usepackage{latexsym}

\usepackage[T1]{fontenc}
\usepackage{amsmath}

\usepackage[utf8]{inputenc}

\usepackage{microtype}

\usepackage{inconsolata}
\usepackage{booktabs}
\usepackage{subfigure}

\usepackage{graphicx}
\newcommand{\second}[1]{\underline{#1}}
\newcommand{\best}[1]{\textbf{#1}}

\usepackage{graphicx}
\usepackage{natbib}
\usepackage{caption}
\usepackage{booktabs}
\usepackage{amsmath}
\usepackage{subfigure}
\usepackage{algorithm}
\usepackage{algorithmicx}
\usepackage{algpseudocode}

\title{TKGB: A Comprehensive Benchmark for Temporal Knowledge Graph Forecasting}

\author{%
  {\small\textbf{Jiasheng Zhang\textsuperscript{1}, Zhengpin Li\textsuperscript{2}, Hao Li\textsuperscript{1},
  Jie Shao\textsuperscript{3}, Hui Li\textsuperscript{1}, Jiangtao Cui\textsuperscript{4}}} \\
  {\small\textsuperscript{1}Xidian University, Xi'an, China\quad
  \textsuperscript{2}Peking University, Beijing, China} \\
  {\small\textsuperscript{3}University of Electronic Science and Technology of China, Chengdu, China} \\
  {\small\textsuperscript{4}Xi'an University of Posts\&Telecommunications, Xi'an, China} \\
  {\footnotesize\texttt{\{zhangjiasheng,hli,cuijt\}@xidian.edu.cn}} \\
  {\footnotesize\texttt{zpli@PKU.edu.cn}\quad\texttt{25031212362@stu.xidian.edu.cn}} \\
  {\footnotesize\texttt{shaojie@uestc.edu.cn}}%
}

\begin{document}
\maketitle
\begin{abstract}
Temporal knowledge graph (TKG) forecasting aims to predict future knowledge from existing knowledge and has advanced remarkably. Yet existing benchmarks inadvertently introduce a shortcut: For a subset of test queries, models can achieve optimal evaluation metrics without learning any temporal dynamics.  We trace this to two pieces of evidence: inherent dataset biases and an overly simplistic task formulation that allows models to exploit biases, revealing fundamental flaws in current evaluation protocols for differentiating true forecasting ability.  While such a flaw motivates us to develop a more comprehensive evaluation, existing datasets rely on fixed history that poorly reflects real-world dynamics, lack semantic context, and fail to explicitly mitigate shortcut learning, preventing thorough model assessment. To address these, we propose TKGB, a comprehensive benchmark for evaluating the capability of TKG forecasting models.  By integrating four complementary tasks with an automatic pipeline for constructing up-to-date and context-enriched TKG datasets with controllable shortcuts, TKGB reveals capability differences and failure modes hidden under conventional evaluation.

\end{abstract}

\section{Introduction}
Knowledge graphs (KGs) are structured representations of real-world knowledge and underpin many applications \cite{zhong2025knowledge, wang2019kgat}. Traditional research \cite{bordes2013translating,wang2014knowledge,lin2015learning} largely focuses on static KGs where facts remain permanently valid. Recently, research has shifted toward temporal knowledge graphs (TKGs), where facts are valid at specific times, e.g., $(Messi, TransfersTo, PSG, 2021/08/10)$ and $(Messi, PlaysFor, PSG,2021/08/10, 2023/07$ $/16)$. Many methods \cite{li2021temporal, jung2021learning} have been proposed, and TKG forecasting, directly adapted from the static KG completion paradigm \cite{bordes2013translating}, has become the standard evaluation protocol: given a subject entity $s$ and relation $r$, predict the missing object entity $o$ at future time $t$. Using these benchmarks, recent methods \cite{wang2024large, cao2025dpcl} report impressive performance. 

\begin{figure}[t]
\centering
\includegraphics[width=0.75\linewidth]{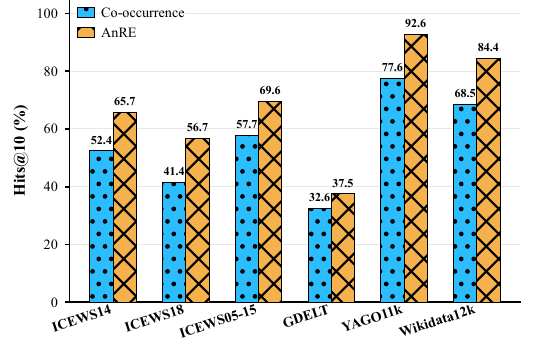}
\caption{Co-occurrence vs. a recent supervised model on existing TKG forecasting benchmarks.}
\label{fig:observations}
\end{figure}

Recently, some studies~\cite{10.24963/ijcai.2024/444} have observed the competitiveness of frequency-based methods on existing TKG benchmarks. As shown in Figure  \ref{fig:observations}, simply replaying frequent entity-relation co-occurrence, with no learnable parameters and ignoring all temporal information, achieves competitive results. It reaches 81.2\% of the SOTA’s average Hits@10, peaking at 86.9\% on the GDELT dataset. Beyond confirming the predictive value of historical frequency, these results motivate us to conduct a closer examination of what existing protocols actually reward for temporal knowledge forecasting evaluation.

We investigate two interacting sources of shortcut susceptibility: an unsuitable evaluation protocol directly copied from static KGs and the repetitive, low-diversity property prevalent in TKGs. Analogous to KG completion, query $(s,r,?,t)$ already reveals that subject $s$ participates in relation $r$ at time $t$, leaving the model only to rank objects. But in TKGs, many \((s,r)\) pairs interact only with a small set of objects, and these interactions have already appeared in history. When this set contains at most $N$ objects and the ground-truth object falls within it, the Hits@$N$ metric can be exploited by memorizing co-occurring objects, without any temporal reasoning. Our analyses show that several existing benchmarks exhibit such low-diversity distributions, and the standard object-ranking evaluation fails to distinguish the capabilities of true temporal inference from simple co-occurrence replaying.

These findings motivate multidimensional evaluation of forecasting. However, existing TKG datasets provide limited support for designing such evaluations. First, they focus on either timestamped events or interval-valid knowledge, without jointly representing occurrence and validity. Their textual annotations also often provide little beyond opaque IDs (e.g., Q1070888 and P463) or abbreviated labels (e.g., OPP REF). Both limit the context needed to understand real-world interactions and knowledge expiration. Second, fixed historical datasets offer limited support for testing on newly emerging facts and changing interaction patterns. For LLM-based models, they also risk pretraining contamination, making forecasting difficult to distinguish from factual recall. These gaps call for broader task coverage, richer context resources, and a reproducible pipeline for constructing and updating datasets with controllable shortcuts.

To address these gaps, we introduce TKGB, a benchmark for multidimensional TKG forecasting evaluation with configurable dataset construction and support for newly collected event windows. Its four tasks assess whether models can anticipate which entities will be active in the future, forecast knowledge with fewer future prompts, retain valid knowledge while suppressing obsolete facts, and recover interactions that simple co-occurrence fails to prioritize. TKGB provides a standardized dataset construction pipeline that automatically retrieves up-to-date real-world events and aligns participants with Wikidata, yielding rich textual descriptions and explicit interval augmentations for studying context validity. It provides configurable domains, temporal granularities, and time windows, with explicit filtering rules to build challenge subsets targeting low-diversity shortcuts. Experiments on TKGB reveal model capability differences missed by conventional benchmarks.

\section{Background}
We denote a temporal knowledge graph as $\mathcal{G}=(\mathcal{V},\mathcal{R},\mathcal{E},\mathcal{T})$, where $\mathcal{V}$ and $\mathcal{R}$ are the entity and relation sets, $\mathcal{T}$ is an ordered time domain, and $\mathcal{E}=\mathcal{E}_{\mathrm{pt}}\cup\mathcal{E}_{\mathrm{int}}$ contains timestamp and interval knowledge (facts). $(s,r,o,t)\in\mathcal{E}_{\mathrm{pt}}$ records a fact occurring at $t$ and $(s,r,o,t_s,t_e)\in\mathcal{E}_{\mathrm{int}}$ records a fact valid over $[t_s,t_e]$, with $t_s\leq t_e$. $s,o\in\mathcal{V}$, $r\in\mathcal{R}$, and $t,t_s,t_e\in\mathcal{T}$. We write $(s,r,o)$ for the underlying triple without temporal annotations.

\textbf{TKG forecasting task.}
TKG forecasting aims to predict future facts from history. Existing benchmarks formulate it as object completion: given a query $(s,r,?,t)$, a model ranks candidate objects and the performance is measured by the rank of the ground-truth object (Hits@$N$ or MRR). As a diagnostic baseline, we introduce a non-parametric co-occurrence strategy: For Hits@$N$, the baseline successively adds objects co-occurring with $(s,r)$, remaining objects co-occurring with $s$ that also occur as objects of $r$ with any subject, and other objects co-occurring with $s$ into candidate sets, until the set size reaches $N$. It uses neither time information nor learnable parameters (see Appendix \ref{app:Co-occurrence}).

\textbf{TKG forecasting datasets.}
Widely used TKG datasets primarily originate from two types of sources. Event monitoring streams like ICEWS and GDELT-Project provide timestamped interactions. Legacy benchmarks (e.g., ICEWS14, ICEWS18, ICEWS05-15, GDELT) are built from fixed temporal snapshots, retaining facts involving frequent entities \cite{garcia2018learning}. General-domain knowledge bases such as YAGO and Wikidata offer interval-validity facts, yielding datasets like YAGO11k and Wikidata12k. To fit timestamp-based evaluation of existing benchmarks, typical approaches expand each interval into repeated facts across its validity period \cite{xu2020temporal}. Recently, TGB 2.0~\cite{gastinger2024tgb} released four large-scale datasets and evaluated models on them using the standard object-ranking protocol. Yet they are still fixed-history datasets without online updating and lack rich semantic contexts. TKGB instead establishes a configurable construction pipeline for context-enriched, refreshable, and shortcut-controlled datasets, complemented by multidimensional evaluation beyond simple object-ranking.

\begin{figure*}[t]
\centering
\subfigure[]{\includegraphics[width=0.48\linewidth]{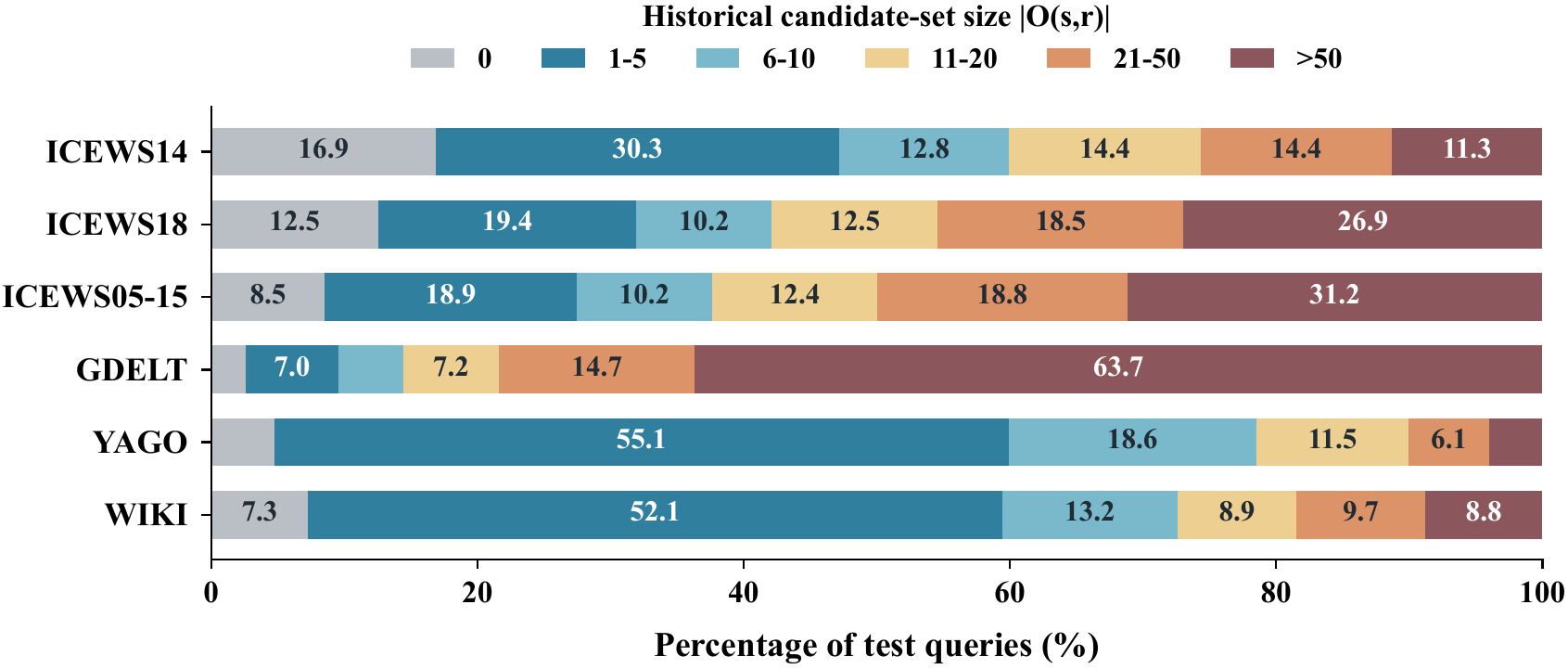}}
\subfigure[]{\includegraphics[width=0.50\linewidth]{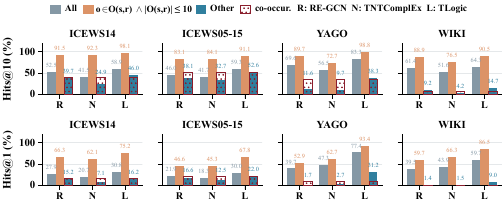}}
\caption{(a) Historical candidate-set size distributions. Each segment shows the percentage of test queries within a size bin of $O{(s,r)}$. (b) Hits@10 and Hits@1 performance across query subsets. All denotes the complete test set. $o\in O{(s,r)} \land |O{(s,r)}|\le10$ identifies queries whose answers belong to a historical set of at most 10 objects. Other denotes the remaining queries.}
\label{fig:observations_2}
\end{figure*}

\textbf{TKG modeling methods.}
TKG models capture temporal information via diverse mechanisms. Embedding-based methods encode timestamps into scoring functions~\cite{goel2020diachronic,ying2024simple}. Graph-based approaches model evolving structure through recurrent updates~\cite{li2021temporal,wu2020temp,han2021learning}. Others exploit historical motifs or logical rules~\cite{liu2022tlogic,li2025infer,chen2025cogntke,tang2025anre}, and language-model-based methods predict via reasoning over historical sequences~\cite{xu2023pre,lee2023temporal,tang-etal-2026-implicit,bai2025g2s}. Recent work~\cite{10.24963/ijcai.2024/444} also observed the competitiveness of frequency heuristics. Yet its analysis still focuses on whether models capture frequency under the object-completion protocol. TKGB instead investigates the protocol itself: whether success under redundant future conditions and the dominance of low-diversity interactions truly reflects forecasting capability, and how dataset construction and task coverage enable broader assessment.

\section{Existing Benchmark Analysis}
Using standard dataset splits, we construct historical candidate sets from training and validation facts. Let \(O(s,r)\) denote the set of objects that have historically co-occurred with a given \((s,r)\) pair. Figure~\ref{fig:observations_2}(a) reveals that 43.1\%, 29.1\%, 73.7\%, and 65.3\% of test queries in ICEWS14, ICEWS05-15, YAGO, and WIKI, respectively, have historical candidate sets containing merely 1–10 entities. A large fraction of test queries thus correspond to low-diversity historical interactions, particularly in YAGO and WIKI. These two datasets expand interval facts into repeated timestamped triples, replicating triples over their periods of validity.

We further investigate queries satisfying \(o\in O(s,r)\) and \(|O(s,r)|\le10\), where the ground-truth answer recurs historically, and the candidate pool remains small. For such queries, simply returning the full historical candidate set for every future timestamp is sufficient to attain strong Hits@10 performance, without requiring the model to infer which interactions will recur at the query time. The upper panels of Figure~\ref{fig:observations_2}(b) demonstrate that existing models yield drastically improved performance on these queries. For instance, TLogic obtains a Hits@10 of 98.1\% versus 46.0\% on ICEWS14, and 98.8\% versus 38.3\% on YAGO. Aggregate metrics are consequently inflated by near-perfect results on this favorable subset, offsetting weaker performance on the remaining queries. As shown in the Other set of ICEWS05-15, simple co-occurrence can achieve 51.20\% Hits@10, compared with 38.1\% for RE-GCN and 32.7\% for TNTComplEx. This issue reveals a critical shortcut: standard object-ranking evaluation rewards naive replay of historical records. It cannot differentiate whether the performance of a model truly stems from time-aware inference or simply from co-occurrence replay.

This performance gap also holds for Hits@1. The lower panels show TLogic reaches 75.2\% versus 16.2\% on ICEWS14, and 93.4\% versus 31.2\% on YAGO. Raising the ranking strictness by switching to Hits@1 therefore fails to erase the performance boost stemming from historically covered answers and compact candidate sets. These comparisons demonstrate that standard object-ranking alone is insufficient to assess model capabilities.

In practice, forecasting systems should proactively alert users to upcoming events by identifying potential participants and their interactions, without requiring users to predefine them. Such systems must also mitigate reliance on outdated associations, refrain from repeatedly predicting familiar interactions that yield little actionable insight, and maintain robustness under pattern shifts. These requirements extend beyond the standard forecasting task, which motivates the introduction of supplementary tasks. Evaluating such real-world capabilities also requires explicit annotations for temporal validity, updatable evaluation windows, and controlled shortcuts. Existing benchmarks do not consistently provide these resources, motivating TKGB's task design and configurable dataset construction.

\section{Dataset Construction Pipeline}
\label{sec:dataset_construction}
TKGB provides a configurable pipeline for constructing semantically enriched and up-to-date TKG datasets from real-world event streams. Under user specifications, raw sources are fetched and iteratively curated via entity grounding, cutoff-aware context augmentation, and subset construction, and released reproducibly. The resulting datasets serve as the backbone for TKGB tasks, supporting evaluation on both full test facts and shortcut-controlled subsets, with semantic descriptions and interval facts as optional auxiliary context. Implementation details are in Appendix~\ref{app:datasets}.

\paragraph{User specification.}
A construction request is
\(
\Omega =
\left(
\mathcal{W}_{\mathrm{tr}},
\mathcal{W}_{\mathrm{val}},
\mathcal{W}_{\mathrm{te}},
g,d,m,\theta
\right),
\)
where $\mathcal{W}_{\mathrm{tr}}$,
$\mathcal{W}_{\mathrm{val}}$, and
$\mathcal{W}_{\mathrm{te}}$ are non-overlapping and chronologically
ordered training, validation, and test windows, respectively.
$g\in\{\mathrm{day},\mathrm{month},\mathrm{year}\}$ specifies the
temporal granularity. $d$ denotes the semantic domain.
$m\in\{\mathrm{offline},\mathrm{online}\}$ specifies the evaluation
mode, and $\theta$ contains the shortcut-filter parameters.
Given $\Omega$, TKGB produces fact splits
\(
\mathcal{D}_{\Omega}
=
\left\{
\mathcal{E}^{\mathrm{tr}},
\mathcal{E}^{\mathrm{val}},
\mathcal{E}^{\mathrm{te}}
\right\},
\)
together with separately versioned entity descriptions and interval
context. These controls support different evaluation settings. The
granularity $g$ ranges from fine-grained event monitoring to
coarse-grained long-horizon forecasting, while $d$
specifies a topical scope such as politics or finance. The offline mode uses predefined chronological splits, whereas the online mode evaluates models on the latest event stream. In both modes, descriptions and interval context are exposed according to the prediction cutoff rather than being assigned unconditionally to every split.


\paragraph{Event candidate Fetching.}
Given a specification \(\Omega\), TKGB retrieves GDELT records within each window. Each record contains two participant mentions, a relation label, and an occurrence time. TKGB normalizes these records into timestamped facts \((s,r,o,t)\). Here $s$ and $o$ are participant mentions, $r$ denotes a relation type, and $t$ follows the required granularity, i.e., YYYY-MM-DD, YYYY-MM, or YYYY for day-, month-, and year-level settings. Raw record streams include generic categories and non-instance expressions that cannot be reliably grounded as entities. We perform an initial concept filter to keep \(\mathcal{E}^*_{cand}=\{(s,r,o,t)\mid s,o\notin \mathcal{B}_{concept}\}\), where \(\mathcal{B}_{concept}\) is a blacklist manually constructed and extended by an LLM, which is shared across splits. $*$ can be selected from $\{tr,val,te\}$.


\paragraph{Wikidata alignment and entity grounding.}
Raw-stream participant mentions are not canonical identifiers and may contain abbreviations or variants, hindering reliable aggregation. TKGB therefore grounds these mentions to Wikidata items, which provide stable identifiers and
structured entity metadata. For each mention $n$, we collect its event context $C(n)
=
\left\{
(s,r,o,t)
\mid s=n \lor o=n
\right\}$ from the whole $\mathcal{E}_{cand}$ and ask the LLM to propose alternative names $\hat{N}(n)=\mathrm{LLM}_{\mathrm{name}}(n,C(n))$. The original mention and the proposed names are then queried against Wikidata. Mentions with no matches are treated as unresolved. For each retrieved candidate item $q$, the LLM then verifies whether the description and entity type of $q$ are semantically compatible with the relations in $\mathcal{C}(n)$ and selects the most compatible item. Events containing an unresolved participant or an item that fails this check are discarded. The LLM is therefore used only to propose search names and validate a retrieved item. It is not allowed to generate a Wikidata identifier directly. Appendix~\ref{app:datasets} reports a human audit of the retained entity annotations, covering entity alignment, domain labels, and descriptions.

\paragraph{Context augmentation.}
After grounding GDELT participants, TKGB provides two types of auxiliary context: entity descriptions and time-interval facts. First, TKGB collects the Wikidata short description for each grounded entity, which summarizes the persistent identity and type of a Wikidata item, e.g., `Lionel Messi' is described as `Argentine football player'. To reduce potential temporal leakage, we directly remove dates and date ranges from these descriptions.

Second, TKGB extracts interval facts from Wikidata claims for each grounded entity, which provide background context. For instance, the event $(\text{Messi},\text{honored}, \text{Best Foreign Player in Ligue 1})$ should occur during Messi's tenure at PSG. TKGB creates a split-specific view of each interval fact. For a split ending at $t_b$, an interval fact $(s,r,o,t_s,t_e)$ satisfying $t_s<t_b<t_e$ is stored in that split as a boundary-truncated version $(s,r,o,t_s,t_b)$. Consequently, no split stores an endpoint occurring after its temporal boundary.

\paragraph{Subset Construction.}
After entity grounding, TKGB builds scenario-specific subsets from domain labels of grounded entities. For an entity $n$, let \(D(n)\) denote its domain labels. TKGB retains facts where \(D(s)\) or \(D(o)\) matches a target domain $d$, and adds induced facts connecting all relevant entities:
\(\mathcal{E}^* =
\{(s,r,o,t) \in \mathcal{E}^*_{cand}\mid D(s)=d\vee D(o)=d\} \cup \{(s,r,o,t) \in \mathcal{E}^*_{cand} \mid s,o\in\mathcal{U}_{d}\},\)
where \(\mathcal{U}_{d}\) is the entity set from the initially selected facts. This procedure constructs coherent domain-specific TKGs while keeping contextual entity interactions. The offline setting directly outputs train/validation/test splits. For the online setting, test data is built via the same pipeline on a fixed snapshot of the latest GDELT stream.

Beyond the full test set, TKGB can provide shortcut-controlled subsets that remove test facts solvable via low-diversity historical preferences. Let \(\mathcal{E}^{\mathrm{tr}}, \mathcal{E}^{\mathrm{val}}, \mathcal{E}^{\mathrm{te}}\) denote the previously derived splits. We define historical candidate sets:
\(
O(s,r) = \left\{ o'\mid(s,r,o',t')\in\mathcal{E}^{\mathrm{tr}} \right\},
O(s) = \left\{ o'\mid(s,r',o',t')\in\mathcal{E}^{\mathrm{tr}} \right\}.\)
A test fact \((s,r,o,t) \in \mathcal{E}^{\mathrm{te}}\) is marked shortcut-prone if
\(o\in O(s,r) \land |O(s,r)|\leq\theta_{sr}\),
or
\(o\in O(s) \land |O(s)|\leq\theta_s\),
with \(\theta_{sr},\theta_s\) as diversity thresholds. Filtering these items yields a challenging test subset. Each release ships datasets, filtering logs, source records, metadata versions, and cryptographic hashes of canonically sorted facts to guarantee reproducibility.

\section{TKGB Tasks}
\paragraph{Entity Activation Estimation.}
Traditional TKG forecasting evaluates predefined queries whose source entity $s$ is guaranteed to have interactions at timestamp $t$. In practical scenarios, however, the system may receive queries of entities actually inactive at $t$. The model
should therefore recognize whether a query entity will be active at $t$ and suppress forecasting for inactive entities. This task evaluates whether a model assigns higher future-interaction scores to active entities than to historically prominent but currently inactive entities.

At each test timestamp $t$, active subjects are defined as \(\mathcal{A}_{t}=\{s\mid(s,r,o,t)\in\mathcal{E}^{te}\}\). We build a hard negative inactive set \(\mathcal{N}_{t}\) using the most frequent historical subjects that have no interactions during $[t-\mathcal{L},t]$. Given historical facts $\mathcal{E}_{t-}$ before $t$, the model scores possible facts for each query $(s,?,?,t)$. We aggregate its top-$K$ fact scores into an entity-level activation score:
\[
\alpha_{\theta}(s,t)=
\frac{1}{K}\sum
\operatorname{TopK}_{(r,o)}
f(s,r,o,t\mid\mathcal{E}_{t-}).
\]

We report ROC-AUC over $\mathcal{A}_{t}\cup\mathcal{N}_{t}$ using activation scores, which measures the model's ability to discriminate active entities from inactive entities.

\begin{figure}[t]
\centering
\includegraphics[width=0.95\linewidth]{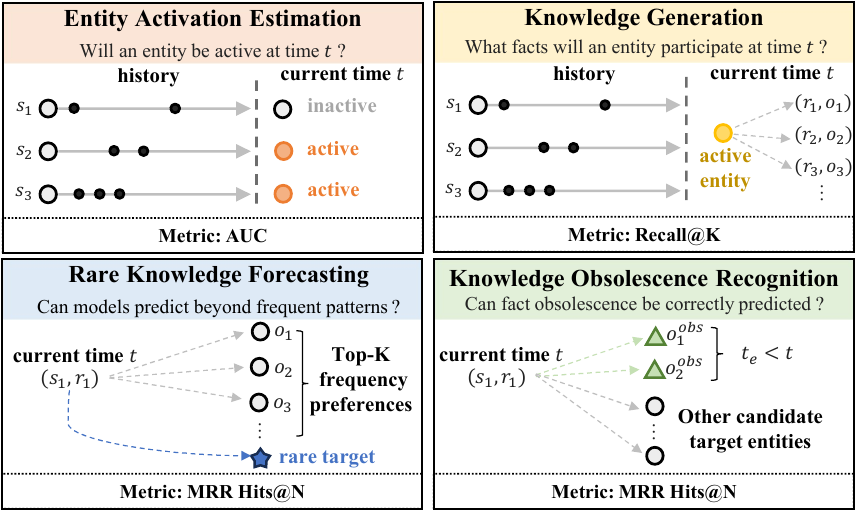}
\caption{TKGB benchmarking tasks.}
\label{fig:benchmarking_tasks}
\end{figure}

\paragraph{Knowledge Generation.}
Traditional TKG forecasting assumes that both the subject and
relation of a future query are known and ranks only candidate
objects. In practical applications, however, a system may need to
start from an entity of interest and predict its future interactions
without being given the relation. Given a query subject $s$, the
model produces ranked relation--object pairs from the joint
candidate space $\mathcal{R}\times\mathcal{V}$:
\[
\hat{\mathcal{G}}_{t}(s)=
\operatorname{TopK}_{(r,o)}
f(s,r,o,t\mid\mathcal{E}_{t-}).
\]

The observed target set is
\(
\mathcal{G}_{t}(s)=
\{(r,o)\mid(s,r,o,t)\in\mathcal{E}^{te}\}.
\)
Since real-world TKGs are incomplete under the open-world
assumption, we use Recall@$K$ to measure the coverage of observed
future facts under different generation budgets.

\paragraph{Knowledge Obsolescence Recognition.}
Temporal knowledge may expire as well as emerge. A forecasting
model should therefore retrieve facts that remain valid at timestamp
$t$ while suppressing facts whose validity has ended. Let $\mathcal{I}$ denote the augmented interval facts. For a query $(s,r,?)$ constructed from an interval fact and the query timestamp $t$, we define the valid object set as
\[
\mathcal{O}^{+}_{t}(s,r)=
\left\{
o\mid
(s,r,o,t_s,t_e)\in\mathcal{I},
\ t_s\leq t\leq t_e
\right\},
\]
and the obsolete object set as
\[
\begin{aligned}
\mathcal{O}^{\mathrm{obs}}_{t}(s,r)
=\bigl\{o' \mid {}&
(s,r,o',t_s,t_e)\in\mathcal{I},\\
& t_e<t,\quad o'\notin O_t^+(s,r)
\bigr\}.
\end{aligned}
\]
Given the history before $t$, the model ranks both valid and obsolete objects within the same candidate space $\mathcal{V}$. We report Valid-Hits@$N$/Obs-Hits@$N$ and Valid-MRR/Obs-MRR as paired metrics. They are interpreted jointly: high valid-object scores together with low obsolete-object scores indicate effective obsolescence recognition, whereas low scores for both indicate generic under-ranking.

\textbf{Rare Knowledge Forecasting.}
Frequency is a useful predictive signal, but models may rely excessively on frequent objects. We therefore evaluate whether model predictions remain effective for rare objects falling outside historical frequency preferences. For each test fact $(s,r,?,t)$, we define the top-$K$ frequent historical objects for $(s,r)$ as $O^{K}(s,r)$, and the top-$K$ frequent historical objects for $s$ as $O^{K}(s)$. The rare test fact subset is then defined as $\mathcal{E}^{te}_{rare}=\{(s,r,o,t)\in\mathcal{E}^{te}\mid o\notin O^{K}(s,r) \wedge o\notin O^{K}(s)\}$. For each query $(s,r,?,t) \in \mathcal{E}^{te}_{rare}$, models rank all candidate objects from $\mathcal{V}$, and we report MRR and Hits@$N$. Metric details are in Appendix~\ref{app:metric}.

\begin{table}[t]
\centering
\small
\setlength{\tabcolsep}{0.38mm}
\begin{tabular}{@{}l|c|c|c|c|c|c|c|c}
\toprule
Data & Ent. & Rel. & T. & Tr. & Val. & Test. & Desc. & Int. \\
\midrule
Pol.-D & 5,873  & 343 & 5.4K & 881K  & 9.3K  & 512K  & 99.8\% & 8.7K  \\
Pol.-M & 5,972  & 349 & 180   & 690K  & 7.1K  & 368K  & 99.8\% & 8.8K  \\
Pol.-Y & 7,466  & 380 & 15    & 467K  & 7.5K  & 233K  & 99.7\% & 11.7K \\
Fin.-D & 8,186  & 361 & 5.4K & 2.00M & 16.2K & 1.95M & 99.9\% & 13.7K \\
Fin.-M & 8,476  & 366 & 180   & 1.50M & 9.7K  & 1.18M & 99.8\% & 14.0K \\
Fin.-Y & 11,226 & 399 & 15    & 925K  & 11.2K & 656K  & 99.7\% & 19.0K \\
Med.-D & 8,274  & 361 & 5.4K & 2.05M & 20.1K & 1.99M & 99.8\% & 13.7K \\
Med.-M & 8,564  & 366 & 180   & 1.54M & 13.2K & 1.22M & 99.7\% & 14.0K \\
Med.-Y & 11,296 & 399 & 15    & 961K  & 14.3K & 683K  & 99.6\% & 19.0K \\
\bottomrule
\end{tabular}
\caption{Dataset statistics. T. denotes timestamps; Desc. denotes the proportion of facts whose two entities both have descriptions. Int. denotes unique interval facts.}
\label{tab:dataset_statistics}
\end{table}

\section{Experiments}
Using our framework, we construct nine datasets spanning diverse domains and temporal granularities. Their statistics and details are summarized in Table~\ref{tab:dataset_statistics} and Appendix~\ref{app:datasets}. We select representative methods with official publicly available implementations from embedding-based, graph-based, rule-based, and LLM-based model families. We evaluate models including DE-SimplE~\cite{goel2020diachronic}, TNT~\cite{DBLP:conf/iclr/LacroixOU20}, TCompoundE~\cite{ying2024simple}, RE-GCN~\cite{li2021temporal}, CognTKE~\cite{chen2025cogntke}, TLogic~\cite{liu2022tlogic}, INFER~\cite{li2025infer}, ExE-LLM~\cite{tang-etal-2026-implicit}, and G2S~\cite{bai2025g2s}, using their official implementations. We use task-aligned training protocols. For relation-conditioned object-ranking tasks, neural baselines retain their original training objectives. For knowledge generation, we use an explicitly documented joint relation–object ranking adaptation. Entity activation reuses the generation checkpoint because its scores are aggregated from generated candidate facts. Interval-enhanced variants are fine-tuned separately. Implementation details are in Appendix~\ref{app:implementations}. More experimental results are in Appendix~\ref{app:more_exp}.

\begin{figure*}[t]
\centering
\includegraphics[width=\linewidth]{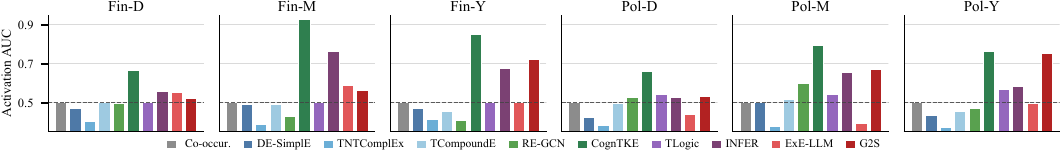}
\caption{Entity activation estimation results.  Each panel reports ROC-AUC for distinguishing active entities from hard inactive entities. Co-occur. denotes the proposed co-occurrence-based scoring strategy. }
\label{fig:activation}
\end{figure*}

\subsection{Entity Activation Estimation}
Figure~\ref{fig:activation} shows that most models achieve near-random performance when distinguishing active subjects from historically frequent hard inactive subjects. CognTKE consistently achieves the highest performance across domains and granularities, which is consistent with its use of recent local structures. In contrast, although RE-GCN also models graph structure, its performance mostly remains close to random, suggesting that global historical structure alone is insufficient for distinguishing local activation states. G2S performs substantially better on year-level datasets, indicating that its semantic signals are more effective for coarse-grained activation patterns. INFER performs better on month-level datasets, also indicating the inability of its rules to capture short-term activations.

\begin{table*}[t]
\centering
\small
\setlength{\tabcolsep}{1mm}
\begin{tabular}{@{}lcccccccccc@{}}
\toprule
Data & Co-occur. & DE & TNT & TCompoundE & RE-GCN & CognTKE & TLogic & INFER & ExE-LLM & G2S \\
\midrule
Fin.-D   & .041/.371 & .000/.018 & .000/.000 & .000/.000 & .029/.252 & .083/.331 & .029/.087 & \best{.354/.795} & .329/.468 & \second{.343/.690} \\
Fin.-M   & .048/.251 & .000/.003 & .000/.002 & .000/.000 & .043/.261 & .014/.177 & .019/.080 & \best{.292/.625} & .177/.253 & \second{.256/.525} \\
Fin.-Y   & .029/.192 & .000/.000 & .010/.010 & .000/.000 & .029/.094 & .032/.155 & .017/.059 & \best{.162/.447} & .031/.071 & \second{.139/.431} \\
Med.-D  & .040/.337 & .000/.003 & .000/.000 & .000/.000 & .028/.296 & .066/.301 & .028/.077 & \second{.377/.668} & .338/.430 & \best{.378/.636} \\
Med.-M  & .031/.263 & .000/.000 & .000/.000 & .000/.000 & .015/.121 & .039/.246 & .006/.061 & \best{.289/.635} & .149/.209 & \second{.219/.537} \\
Med.-Y  & .040/.163 & .000/.000 & .013/.024 & .000/.013 & .004/.047 & .018/.105 & .011/.043 & \best{.151/.425} & .043/.079 & \second{.133/.408} \\
Pol.-D   & .041/.369 & .000/.000 & .000/.000 & .000/.000 & .101/.298 & .074/.312 & .030/.163 & \best{.409/.730} & .305/.402 & \second{.346/.648} \\
Pol.-M   & .053/.249 & .000/.000 & .000/.000 & .000/.000 & .087/.203 & .036/.230 & .021/.108 & \best{.284/.540} & .110/.170 & \second{.216/.487} \\
Pol.-Y   & .026/.190 & .000/.001 & .000/.000 & .000/.000 & .062/.142 & .022/.099 & .010/.077 & \second{.146/.404} & .030/.088 & \best{.152/.430} \\
\bottomrule
\end{tabular}
\caption{Knowledge generation results. Each cell reports Recall@100 and Recall@1000. Best and second-best results are marked by bold and underline, ranked by Recall@100.}
\label{tab:generation-main}
\end{table*}

\begin{figure}[t]
\centering
\includegraphics[width=\linewidth]{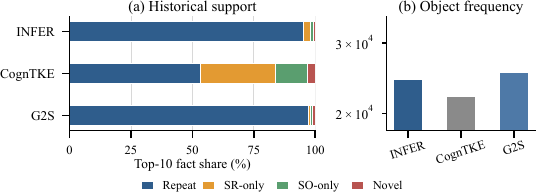}
\caption{Left: statistics of the top-10 generated facts by different models on the Politics-Day dataset. Right: mean historical frequency of the generated object entities.}
\label{fig:top10_fact_diagnostic}
\end{figure}

\begin{table}[t]
\centering
\small
\setlength{\tabcolsep}{0.6mm}
\begin{tabular}{@{}lccccc@{}}
\toprule
Dataset & TNT & RE-GCN & CognTKE & INFER & ExE-LLM \\
\midrule
Fin.-D & .308/.392 & .070/.008 & .051/.006 & .324/.306 & .081/.047 \\
Fin.-M & .331/.277 & .045/.045 & .095/.006 & .385/.255 & .127/.039 \\
Fin.-Y & .229/.223 & .007/.002 & .120/.013 & .316/.324 & .107/.044 \\
Pol.-D & .634/.726 & .006/.000 & .030/.004 & .422/.342 & .047/.015 \\
Pol.-M & .402/.634 & .001/.000 & .061/.007 & .322/.360 & .068/.015 \\
Pol.-Y & .323/.420 & .001/.000 & .148/.063 & .269/.308 & .042/.018 \\
\bottomrule
\end{tabular}
\caption{Knowledge obsolescence recognition results. Each cell is Hits@10 for valid and obsolete objects.}
\label{tab:obsolescence-suppression-representative}
\end{table}

\subsection{Knowledge Generation}
Table~\ref{tab:generation-main} demonstrates that joint relation-object generation remains challenging: most models fall short of the co-occurrence baseline when ranking over the expanded \((r,o)\) candidate space. Notably, several embedding-based models yield near-zero Recall@K. As all target relations and objects exist in the training vocabulary and all models share the identical candidate pool, such near-zero scores reveal that these models struggle to rank ground-truth pairs within the top-$K$ of this large joint search space. INFER attains the best recall across most datasets, benefiting from rule-constrained \((r,o)\) hypotheses that shrink the search space. G2S and ExE-LLM also surpass most baselines, validating the value of modeling textual dependencies.

In Figure~\ref{fig:top10_fact_diagnostic}, Repeat, SR-only, and SO-only label predictions \((s,r',o',t)\) where the triple \((s,r',o')\), pair \((s,r')\), or pair \((s,o')\) has appeared historically, while Novel marks candidates lacking such historical evidence. G2S and INFER predominantly generate repeats with high-frequency objects, reflecting reliance on historical replay. CognTKE generates more non-repetition candidates, aligning with its high-order correlation modeling. Nevertheless, its lower recall implies a trade-off between replaying observed facts and producing diverse candidates.

\subsection{Knowledge Obsolescence Recognition}
After incorporating interval facts into training, Table~\ref{tab:obsolescence-suppression-representative} jointly reports Valid-Hits@10 and Obs-Hits@10. TNT retrieves relatively more valid interval objects, but it also ranks many obsolete objects within the top 10. This pattern is consistent with stable relation-entity
preference modeling that retains historical knowledge without reliably distinguishing its current validity. INFER exhibits a similar pattern. In contrast, RE-GCN, CognTKE, and ExE-LLM obtain lower scores for both Obs-Hits@10 and Valid-Hits@10, indicating their general under-ranking of interval objects with sparser graph structure. Overall, these observations reveal the limited ability of existing models to distinguish currently valid from expired interval knowledge. Effective obsolescence recognition requires combining stable relation-entity evidence to model long-term validity with local context to suppress expired ones.

\begin{table}[t]
\centering
\small
\setlength{\tabcolsep}{0.6mm}
\begin{tabular}{lccccc}
\toprule
Dataset & Co-occur. & RE-GCN & CognTKE & INFER & G2S \\
\midrule
Fin.-D & .248/.479 & .223/.436 & \best{.492/.761} & \second{.396/.573} & .154/.323 \\
Fin.-M & \second{.175/.360} & .158/.344 & .157/.331 & \best{.257/.431} & .125/.278 \\
Fin.-Y & .144/.270 & .063/.145 & \second{.150/.305} & \best{.160/.314} & .093/.235 \\
Pol.-D & .233/.448 & .199/.392 & \best{.468/.723} & \second{.312/.547} & .162/.360 \\
Pol.-M & .189/.378 & .169/.371 & \second{.203/.404} & \best{.242/.456} & .117/.290 \\
Pol.-Y & .149/.287 & .049/.077 & \second{.154/.318} & \best{.164/.335} & .086/.212 \\
\bottomrule
\end{tabular}%
\caption{Standard TKG forecasting results. Each cell reports Hits@10 after and before shortcut filtering.}
\label{tab:exp1-transductive-representative}
\end{table}

\begin{table*}[t]
\centering
\small
\setlength{\tabcolsep}{1mm}
\begin{tabular}{@{}lcccccccccc@{}}
\toprule
Dataset & Co-occur. & DE & TNT & TCompoundE & RE-GCN & CognTKE & TLogic & INFER & ExE-LLM & G2S \\
\midrule
Fin.-D   & .031/.035 & .034/.053 & .033/.051 & .008/.013 & .040/.080 & \best{.441/.681} & .073/.136 & \second{.121/.193} & .049/.135 & .042/.061 \\
Fin.-M   & .030/.044 & .035/.066 & .035/.072 & .004/.005 & .044/.095 & \second{.060/.125} & .046/.112 & \best{.089/.143} & .046/.097 & .047/.075 \\
Fin.-Y   & .042/.080 & .048/.100 & .042/.092 & .005/.006 & .040/.087 & .049/.103 & \best{.062/.134} & \second{.055/.113} & .036/.095 & .038/.060 \\
Med.-D  & .030/.036 & .035/.053 & .034/.054 & .006/.009 & .042/.088 & \best{.449/.690} & .067/.134 & \second{.115/.184} & .045/.105 & .040/.075 \\
Med.-M  & .030/.044 & .035/.067 & .034/.065 & .003/.003 & .043/.099 & \second{.058/.126} & .049/.117 & \best{.085/.135} & .036/.083 & .044/.067 \\
Med.-Y  & .040/.076 & .046/.095 & .040/.086 & .005/.006 & .046/.109 & .045/.100 & \best{.057/.124} & \second{.054/.111} & .041/.096 & .042/.063 \\
Pol.-D   & .030/.037 & .036/.065 & .035/.064 & .005/.007 & .046/.114 & \best{.428/.642} & .053/.118 & \second{.087/.191} & .036/.075 & .047/.071 \\
Pol.-M   & .030/.051 & .034/.067 & .035/.072 & .004/.005 & .047/.105 & \best{.098/.185} & .047/.108 & \second{.071/.152} & .027/.064 & .034/.064 \\
Pol.-Y   & .034/.065 & .040/.082 & .037/.078 & .008/.014 & .022/.037 & .039/.084 & \best{.054/.115} & \second{.049/.105} & .043/.104 & .037/.054 \\
\bottomrule
\end{tabular}
\caption{Rare knowledge forecasting results. Each cell reports MRR and Hits@10 under time-filtered ranking.}
\label{tab:exp1-hard-main}
\end{table*}

\begin{table*}[t]
\centering
\small
\setlength{\tabcolsep}{0.68mm}
\begin{tabular}{@{}lccccccccc@{}}
\toprule
Dataset & Co-occur. & DE & TNT & RE-GCN & CognTKE & TLogic & INFER & ExE-LLM & G2S \\
\midrule
Fin.-D & 0.029/0.264 & 0.000/0.008 & 0.000/0.000 & 0.081/0.248 & 0.070/0.256 & 0.034/0.137 & \best{0.197/0.515} & 0.142/0.294 & \second{0.196/0.542} \\
Fin.-M & 0.026/0.211 & 0.000/0.003 & 0.000/0.002 & 0.021/0.184 & 0.027/0.182 & 0.020/0.096 & \second{0.186/0.436} & 0.131/0.228 & \best{0.200/0.510} \\
\midrule
Med.-D & 0.065/0.297 & 0.000/0.004 & 0.000/0.000 & 0.051/0.222 & 0.052/0.310 & 0.026/0.109 & \second{0.250/0.606} & 0.244/0.365 & \best{0.322/0.617} \\
Med.-M & 0.026/0.263 & 0.000/0.000 & 0.000/0.000 & 0.008/0.081 & 0.050/0.258 & 0.019/0.090 & \second{0.235/0.517} & 0.149/0.251 & \best{0.249/0.571} \\
\midrule
Pol.-D & 0.013/0.238 & 0.000/0.003 & 0.000/0.000 & 0.047/0.198 & 0.048/0.266 & 0.002/0.122 & \second{0.175/0.381} & 0.129/0.203 & \best{0.196/0.464} \\
Pol.-M & 0.027/0.138 & 0.000/0.000 & 0.000/0.000 & 0.003/0.004 & 0.045/0.177 & 0.008/0.128 & \second{0.122/0.308} & 0.043/0.074 & \best{0.160/0.404} \\
\midrule
Avg. & 0.031/0.235 & 0.000/0.003 & 0.000/0.000 & 0.035/0.156 & 0.049/0.241 & 0.018/0.114 & \second{0.194/0.461} & 0.140/0.236 & \best{0.221/0.518} \\
\bottomrule
\end{tabular}
\caption{Online knowledge generation results. Each cell reports Recall@100 and Recall@1000. Best and second-best results are marked in bold and underline, ranked by Recall@100.}
\label{tab:online-generation}
\end{table*}

\subsection{Rare Knowledge Forecasting.}
Table~\ref{tab:exp1-transductive-representative} shows that shortcut-controlled filtering reduces the performance of the baselines on the same datasets. Nevertheless, co-occurrence still outperforms several learning-based baselines, indicating that historical repetition remains a useful signal that these baselines do not exploit effectively. Their substantial degradation on the rare-fact subsets in Table~\ref{tab:exp1-hard-main} further demonstrates limited robustness to long-tail interactions. CognTKE achieves the best overall performance, which is consistent with its high-order structure modeling capturing indirect correlations beyond dominant frequencies. This advantage diminishes at coarser granularities, where the aggregation of more events into each timestamp impairs precise path retrieval. Rule-based methods exhibit more stable performance, because of their weaker dependence on dense neighborhoods.

\begin{figure}[t]
\centering
\includegraphics[width=\linewidth]{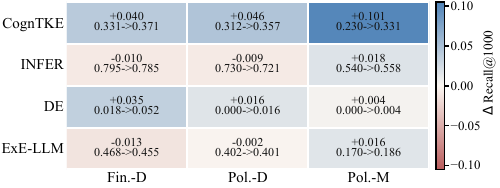}
\caption{Each cell reports knowledge generation performance before and after integrating interval facts.}
\label{fig:interval_context}
\end{figure}

\subsection{Context Effectiveness}
Figure~\ref{fig:interval_context} shows that interval context improves generation performance in most settings. CognTKE benefits most consistently because interval facts can provide additional paths for candidate retrieval. In contrast, ExE-LLM and INFER show less consistent gains. The rules learned by existing methods lack interval knowledge modeling, while the LLM-based models fail to effectively use interval facts as background context. Overall, these results motivate architectures that jointly learn timestamped and interval patterns.

\subsection{Online Testing}
We further create online test sets for each dataset. Each test set is constructed from GDELT events between August 28 and September 27, 2026, with the corresponding domain filters. Online testing uses the corresponding offline task-specific training checkpoints without further parameter updates. Facts between the offline training cutoff and the online test window are available as historical context without parameter updates. We adopt an inductive protocol, with new entities initialized with model-specific strategies if possible, or randomly otherwise. Dataset statistics are reported in Appendix~\ref{app:datasets}. Table~\ref{tab:online-generation} shows that online generation is consistently harder. Co-occurrence decreases from 0.042/0.307 to 0.031/0.235 in Recall@100/Recall@1000, indicating weaker predictability of repetition under online shifts. G2S achieves the best average performance, suggesting that semantic composition transfers better.

\section{Conclusion}
This paper systematically investigates limitations of existing TKG benchmarks that hinder reliable model evaluation. We propose TKGB, a configurable benchmarking framework comprising an automatic dataset construction pipeline and four complementary tasks. Experiments reveal distinct strengths and limitations of existing models.

\section{Limitations}
First, we use GDELT as the sole event source because it combines freely accessible historical archives with frequent, ongoing updates, supporting reproducible and refreshable benchmark construction. It is among the few publicly accessible global event-data resources offering both historical archives and near-real-time updates. However, reliance on a single source may introduce source-specific biases, limiting the generalizability of our findings to other domains or event-extraction pipelines. Second, the shortcut-controlled subsets target explicitly defined repetition and low-diversity patterns, but do not eliminate every possible shortcut. Finally, the four tasks evaluate selected dimensions of forecasting reliability, failing to cover other capabilities of TKG forecasting models. Future work can extend TKGB with additional event sources, open-source construction alternatives, other shortcut diagnostics, and evaluation dimensions.

\bibliography{custom}

\appendix

\clearpage

\section{Co-occurrence strategy}
\label{app:Co-occurrence}
Inspired by co-occurrence counting, which is a widely used heuristic in traditional graph analysis, such as co-neighbors and triangles, we propose a simple TKG forecasting approach based on entity-relation co-occurrence. Specifically, to answer a query $(s,r,?,t)$ and retrieve the top-$N$ most likely candidate objects, the proposed approach first retrieves all entities that have previously served as the object of relation $r$ given the subject $s$, i.e., in facts of the form $(s, r, o)$, and ranks them by their occurrence frequency. If the number of candidates is less than $N$, the method then retrieves entities that have served as the object of subject $s$ in any quadruple, while also having interacted with $r$ (though not necessarily with $s$), and sorts them again by frequency. If the candidate set remains smaller than $N$, it finally includes all other entities that have interacted with $s$ in any relation, ordered by their overall co-occurrence frequency with $s$. Note that this approach is inherently unlearnable and does not incorporate any temporal information.

The implementation relies on three dictionaries preserving co-occurrence frequencies among entities and relations. The first is $e\_2\_e$ which preserves the co-occurrence frequency between two entities. For each entity pair $(s,o)$, their co-occurrence frequency can be defined as $e\_2\_e[s][o] = |\{(r',t') | (s,r',o,t') \in \mathcal{E}^{tr} \}|$. The second is $r\_2\_e$ which preserves the co-occurrence frequency between an entity and a relation. For each entity-relation pair $(r,o)$, their frequency can be defined as $r\_2\_e[r][o] = |\{(s',t') | (s',r,o,t') \in \mathcal{E}^{tr} \}|$. The third is $er\_2\_e$ which preserves the co-occurrence frequency between the subject entity-relation pairs and the object entities. For each pair $(s,r)$ and a candidate object entity $o$, the co-occurrence frequency can be defined as $er\_2\_e[(s,r)][o] = |\{t' | (s,r,o,t') \in \mathcal{E}^{tr} \}|$. Using these dictionaries, we design the co-occurrence strategy to calculate $Hits@10$ for the knowledge forecasting task as shown in Algorithm \ref{alg:time_unkown_model}.

\begin{algorithm}[t]
\caption{Co-occurrence Strategy}
\begin{algorithmic}[1] 
    \State \textbf{Initialize} $total\_num = 0$, $hit\_count = 0$, $hits = 10$
    \State \textbf{Input} $e\_2\_e$, $r\_2\_e$, $er\_2\_e$
    \State \textbf{Output} $\frac{hit\_count}{total\_num}$
    \For{$f \in \mathcal{E}^{te}$} 
        \State Extract subject $s$, relation $r$, object $o$ from f
        \State $total\_num = total\_num + 1$
        \State $candidate\_o\_set = \emptyset$
        
        \If{$(s,r) \in \text{key}(er\_2\_e)$}
            \State $c = [(o', er\_2\_e[(s,r)][o']) \mid o' \in \text{key}(er\_2\_e[(s,r)])]$
            \State Sort $c$ in descending order by the second element
            \For{$(o', val) \in c$}
                \If{$|candidate\_o\_set| < hits$}
                    \State $candidate\_o\_set = candidate\_o\_set \cup \{o'\}$
                \EndIf
            \EndFor
        \EndIf
        
        \State $c = [(o', e\_2\_e[s][o']) \mid o' \in \text{key}(e\_2\_e[s])]$
        \State Sort $c$ in descending order by the second element
        \For{$(o', val) \in c$}
            \If{$|candidate\_o\_set| < hits$ AND $o' \in r\_2\_e[r]$}
                \State $candidate\_o\_set = candidate\_o\_set \cup \{o'\}$
            \EndIf
        \EndFor
        
        \If{$|candidate\_o\_set| < hits$}
            \For{$(o', val) \in c$}
                \If{$|candidate\_o\_set| < hits$}
                    \State $candidate\_o\_set = candidate\_o\_set \cup \{o'\}$
                \EndIf
            \EndFor
        \EndIf
        
        
        \If{$o \in candidate\_o\_set$}
            \State $hit\_count = hit\_count + 1$
        \EndIf
    \EndFor
\end{algorithmic}
\label{alg:time_unkown_model}
\end{algorithm}

\section{Dataset Details}
\label{app:datasets}
TKGB releases each dataset as a self-contained benchmark package. Each package contains timestamp event facts, interval facts, entity metadata, relation metadata, configurations, and hash files. Specifically, facts are stored in \texttt{facts\_train.csv}, \texttt{facts\_valid.csv}, and \texttt{facts\_test.csv}; temporal interval knowledge is stored in \texttt{intervals\_*.csv}; entity and relation metadata are stored in \texttt{entities.csv} and \texttt{relations.csv}. The automatically released \texttt{dataset\_config.json} and hash files make each dataset version traceable and reproducible. We detail the construction and characteristics of the nine datasets used in the experiments as follows.

\textbf{Construction Details.}
TKGB constructs nine datasets across three domains and three granularity levels. For the Politics domain, data are extracted from 1979-01-01 to 1994-01-06, with the period 1979-01-01 to 1989-02-01 as the training window, and the remaining interval as the test window. The Finance and Media domains adopt data from 1989-01-01 to 2004-01-06, following the identical train-validation-test split scheme: 1989-01-01 to 1999-02-01 for training. This difference in time window selection is motivated by the event density characteristics of GDELT: political event streams exhibit sufficient density in early GDELT records, while Finance and Media events require a later time range to achieve stable density. Note that entity grounding is performed as offline identity curation using the full candidate stream. This context is used only to resolve entity identities and is not supplied to forecasting models. Model-visible descriptions and interval facts follow a separate cutoff policy.

Domain labels are assigned from Wikidata entity types. Each domain dataset keeps domain seed entities together with necessary contextual entities and induced interactions, so the resulting graph represents a coherent scenario rather than a keyword-filtered ontology slice. For each domain, TKGB provides day-, month-, and year-level versions generated from the same raw event resources. The validation split is taken from the tail of the historical window, and the test split follows the historical window chronologically. All datasets use the same de-biasing parameters with $\lambda_{sr}=10$ and $\lambda_s=10$. This keeps the construction protocol consistent across datasets while allowing controlled comparison of temporal resolution effects. We use GPT-4o for name proposal and entity–relation compatibility checking. The model identifier, invocation dates, generation settings, retry policy, and response-parsing rules are recorded in the released construction configuration. The prompts are provided below.

\noindent\textbf{Prompt 1: Entity Name Resolution.}
\begin{quote}
Given a noisy GDELT entity name and several observed interaction facts, infer the standard English name or likely Wikidata search names for that entity. Do not return a Wikidata QID. Return compact JSON only with \texttt{canonical\_name}, \texttt{candidate\_names}, \texttt{entity\_type}, \texttt{confidence} as a number from 0 to 1, and \texttt{rationale}.

\medskip
\textbf{GDELT entity name:} \texttt{\{entity\_name\}}

\textbf{Observed interaction facts:}

\texttt{\{context\_facts\}}
\end{quote}

\noindent\textbf{Prompt 2: Entity-Relation Compatibility Validation.}
\begin{quote}
Decide whether the matched Wikidata entity is a concrete instance that can plausibly participate in the observed GDELT relations. Reject conceptual names, creative works, domains, numbers, and other non-actors when the observed relation would be implausible. Return compact JSON only with \texttt{is\_valid} as \texttt{true} or \texttt{false} and \texttt{reason}. Do not include a confidence or score field.

\medskip
\textbf{GDELT entity name:} \texttt{\{entity\_name\}}

\textbf{Matched Wikidata QID:} \texttt{\{qid\}}

\textbf{Matched Wikidata label:} \texttt{\{wikidata\_label\}}

\textbf{Matched Wikidata description:} \texttt{\{wikidata\_description\}}

\textbf{Observed GDELT interaction facts:}

\texttt{\{context\_facts\}}
\end{quote}

\begin{figure}[t]
\centering
\includegraphics[width=0.9\linewidth]{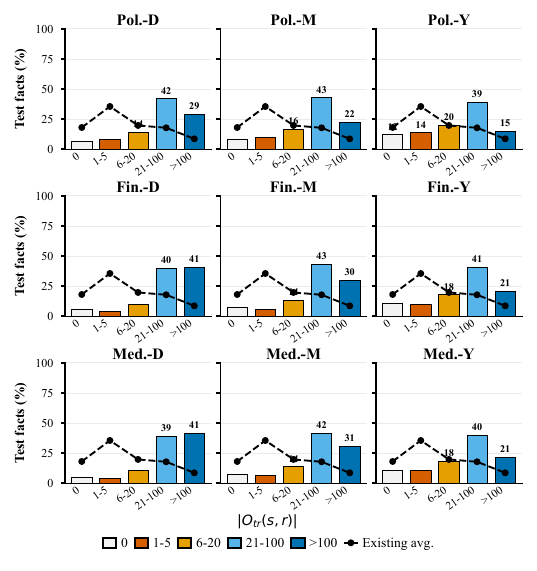}
\caption{Diversity distributions of TKGB datasets. 
Each subplot shows the distribution of historical subject-relation preference set sizes $|O(s,r)|$ over test facts. 
Bars denote the proportion of test facts falling into each preference-size bin, and the black dashed curve denotes the average distribution of existing TKG benchmarks.}
\label{fig:preference_diversity_distribution}
\end{figure}

\begin{table}[t]
\centering
\small
\setlength{\tabcolsep}{3pt}
\renewcommand{\arraystretch}{1.1}
\resizebox{\linewidth}{!}{
\begin{tabular}{l|c|c|c|c}
\toprule
Domain &
\shortstack{Audited\\entities} &
\shortstack{Entity alignment\\accuracy} &
\shortstack{Domain label\\accuracy} &
\shortstack{Description\\accuracy} \\
\midrule
Politics & 2000 & 0.998 & 1.0 & 0.986 \\
Finance  & 2000 & 0.997 & 0.995 & 0.989 \\
Media    & 2000 & 0.998 & 1.0 & 0.992 \\
\midrule
Overall  & 6000 & 0.997 & 0.998 & 0.989 \\
\bottomrule
\end{tabular}
}
\caption{Human audit of the entity-level annotation quality of the constructed datasets.}
\label{tab:human_audit}
\end{table}

\begin{table}[t]
\centering
\small
\setlength{\tabcolsep}{0.9mm}
\begin{tabular}{@{}l|c|c|c|c|c@{}}
\toprule
Data & Ent. & Rel. & Trans. & Ind. & Desc. \\
\midrule
Pol.-D online testing & 5,873 & 207 & 64.2K & 4.7K & 99.8\% \\
Pol.-M online testing & 5,972 & 207 & 46.7K & 4.0K & 99.8\% \\
Fin.-D online testing & 8,186 & 203 & 56.1K & 3.6K & 100.0\% \\
Fin.-M online testing & 8,476 & 203 & 38.4K & 2.9K & 100.0\% \\
Med.-D online testing & 8,274 & 205 & 59.4K & 3.0K & 100.0\% \\
Med.-M online testing & 8,564 & 205 & 41.0K & 2.3K & 100.0\% \\
\bottomrule
\end{tabular}
\caption{Online testing set statistics. Trans. denotes transductive facts whose entities are both seen during training. Ind. denotes candidate online event
facts involving at least one entity unseen in offline training.
Desc. denotes the proportion of Test. facts whose two entities both
have textual descriptions.}
\label{tab:online_dataset_statistics}
\end{table}

\textbf{Dataset Analysis. }
We further analyze whether the constructed datasets reduce the shortcut patterns identified in existing benchmarks. Figure~\ref{fig:preference_diversity_distribution} gives a finer-grained view of the nine TKGB datasets. For each test fact, we compute the size of its historical subject-relation preference set $|O(s,r)|$. A small value means that the answer can be selected from only a few historically observed objects, which makes the query more vulnerable to memorization. Across the nine datasets, most test facts are shifted away from the lowest-diversity bins, while the black dashed curve shows that existing benchmarks have a higher concentration in these bins. This result shows that TKGB provides a more diverse distribution for TKG model evaluation.

To assess the quality of automatically constructed datasets, we randomly sample 2000 entity records from each domain and manually evaluate three aspects. Entity alignment is correct when the matched Wikidata entity corresponds to the original GDELT entity given its name and event context. Domain labels are checked against human knowledge. Descriptions are correct when they refer to the matched entity and the description contains no temporal information leakage. Accuracy is calculated as the proportion of records verified as correct among all audited records.  Table~\ref{tab:human_audit} reports domain-specific and pooled accuracy. Since the day-, month-, and year-level versions share entity metadata, each sampled record is audited only once. These audit results provide quality evidence for our automatic dataset construction pipeline. We also release the finalized entity mappings and filtering decisions together with the dataset splits, enabling evaluation without repeating LLM-assisted alignment. In Table~\ref{tab:online_dataset_statistics}, we report the statistics of the online test sets.

\section{Metric Details}
\label{app:metric}
All metrics use historical information available before the query timestamp. Candidate vocabularies, prediction directions, and tie-breaking rules are fixed within each experiment.

\textbf{Entity Activation Estimation.}
At each test timestamp $t$, active subjects $\mathcal{A}_t$ are
positive samples, while historically prominent but inactive subjects
$\mathcal{N}_t$ form the hard-negative population. Given the
activation score $\alpha_{\theta}(s,t)$ defined in the main paper,
we compute
\begin{equation}
\begin{aligned}
\mathrm{AUC}_t
&=
\frac{1}{|\mathcal{A}_t||\mathcal{N}_t|}
\sum_{s^+\in\mathcal{A}_t}
\sum_{s^-\in\mathcal{N}_t}
\Bigl(
\mathbf{1}[\alpha_{\theta}(s^+,t)>\\
                 &\alpha_{\theta}(s^-,t)] 
+
\frac{1}{2}
\mathbf{1}[\alpha_{\theta}(s^+,t)=
                 \alpha_{\theta}(s^-,t)]
\Bigr).
\end{aligned}
\end{equation}

The reported ROC-AUC is the macro-average over timestamps with positive and negative sets:
\begin{equation}
\mathrm{ROC\text{-}AUC}
=
\frac{1}{|\mathcal{T}_{\mathrm{eval}}|}
\sum_{t\in\mathcal{T}_{\mathrm{eval}}}\mathrm{AUC}_t,
\end{equation}
where comparisons are made at the same timestamp. Higher ROC-AUC indicates better separation of active subjects from the inactive subjects.

\textbf{Knowledge Generation.}
For a query $(s,?,?,t)$, the model ranks distinct relation-object pairs in $\mathcal{R}\times\mathcal{V}$ without receiving the ground-truth future relation. Let $\hat{\mathcal{G}}^K_t(s)$ denote the top-$K$ predicted pairs and
\(
\mathcal{G}_t(s)
=
\{(r,o)\mid(s,r,o,t)\in\mathcal{E}^{te}\}
\)
denote the observed target set. We evaluate queries in
$\mathcal{Q}_{\mathrm{gen}}
=\{(s,t):|\mathcal{G}_t(s)|>0\}$ using
\begin{equation}
\mathrm{Recall@}K
=
\frac{1}{|\mathcal{Q}_{\mathrm{gen}}|}
\sum_{(s,t)\in\mathcal{Q}_{\mathrm{gen}}}
\frac{
|\hat{\mathcal{G}}^K_t(s)\cap\mathcal{G}_t(s)|
}{
|\mathcal{G}_t(s)|
}.
\end{equation}

For knowledge generation, trainable neural baselines combine their original training objective with joint relation–object ranking, explicitly comparing cross-relation candidates. The resulting checkpoint is also used for entity activation. Rule-based and training-free methods retain their native procedures.

\textbf{Knowledge Obsolescence Recognition.}
For a query $(s,r,?,t)$, we use the valid object set
$\mathcal{O}^{+}_{t}(s,r)$ and obsolete object set
$\mathcal{O}^{\mathrm{obs}}_{t}(s,r)$ defined in the main paper.
Both are evaluated within the same candidate space $\mathcal{V}$.
Let $\mathrm{rank}_{s,r,t}(o)$ denote the rank of object $o$
according to the model's forecasting score. For either nonempty target set
$\mathcal{S}\in
\{\mathcal{O}^{+}_{t}(s,r),
  \mathcal{O}^{\mathrm{obs}}_{t}(s,r)\}$,
the query-level metrics are
\begin{equation}
\mathrm{MRR}
=
\frac{1}{|\mathcal{S}|}
\sum_{o\in\mathcal{S}}
\frac{1}{\mathrm{rank}_{s,r,t}(o)},
\end{equation}
and
\begin{equation}
\mathrm{Hits@}N
=
\frac{1}{|\mathcal{S}|}
\sum_{o\in\mathcal{S}}
\mathbf{1}[\mathrm{rank}_{s,r,t}(o)\le N].
\end{equation}

Applying these definitions to valid and obsolete objects yields
Valid-MRR/Obs-MRR and Valid-Hits@$N$/Obs-Hits@$N$, respectively.
The paired metrics are interpreted jointly: higher valid-object
metrics together with lower obsolete-object metrics indicate
effective obsolescence recognition.

\textbf{Rare Knowledge Forecasting.}
Rare knowledge forecasting keeps the standard object entity prediction interface but evaluates only the subset $\mathcal{E}^{te}_{rare}$ defined above. For each query $(s,r,?,t)\in\mathcal{E}^{te}_{rare}$, the model ranks all candidate objects $o'\in\mathcal{V}$, and we report MRR and Hits@$N$. Let $\mathrm{rank}(o)$ denote the rank of the ground-truth object after filtering other true facts at timestamp $t$. Then
\begin{equation}
\mathrm{MRR}
=
\frac{1}{|\mathcal{E}^{te}_{rare}|}
\sum_{(s,r,o,t)\in\mathcal{E}^{te}_{rare}}
\frac{1}{\mathrm{rank}(o)},
\end{equation}
and
\begin{equation}
\mathrm{Hits@}N
=
\frac{1}{|\mathcal{E}^{te}_{rare}|}
\sum_{(s,r,o,t)\in\mathcal{E}^{te}_{rare}}
\mathbf{1}[\mathrm{rank}(o)\le N].
\end{equation}

These metrics measure whether models can forecast outside high-frequency historical sets.

\section{Implementation Details}
\label{app:implementations}

\textbf{Baseline Implementation.}
For trainable baselines, unless otherwise specified by the original implementation, we train each model for at most $500$ epochs with an early-stopping patience of $5$. We follow the officially released hyperparameters except for the settings and adaptations specified below. Candidate entities and relations are drawn from the train-visible vocabularies. For embedding-based models, we set the batch size to $1024$ and the embedding dimension to $100$; DE-SimplE uses $64$ negative samples per positive fact, while TNT and TCompoundE use LCWA-style training over train-visible entity candidates with rank $32$. For RE-GCN, we use $2$ recurrent graph convolution layers, hidden dimension $100$, training history length $10$, and test history length $20$. For CognTKE, we use TRED-GNN with $2$ layers, hidden dimension $100$, window size $15$, global history window size $20$, time embedding dimension $16$, and randomly sample at most $128$ neighbors per node per layer during training. For TLogic, we use its official temporal rule learning and rule application pipeline. For INFER, we use rule lengths $\{1,2,3\}$, $100$ random walks, $100$ acyclic walks, rule threshold $0.01$, and an auxiliary KBC module with rank $64$. For LLM-based baselines, we set the history length to $20$ and retrieve $10{,}000$ candidates. ExE-LLM further uses $k_{\mathrm{time}}=k_{\mathrm{struc}}=k_{\mathrm{sem}}=50$, $p=q=0.2$, and $\alpha=\beta=0.5$. We use GPT-4o as the backbone.

For relation-conditioned object-ranking tasks, trainable neural baselines retain their original training objectives. For knowledge generation, we combine each model’s original objective with a joint relation–object ranking objective using equal weights:
\[
\mathcal{L}_{\mathrm{gen}}
=0.5\mathcal{L}_{\mathrm{orig}}
+0.5\mathcal{L}_{\mathrm{joint}}.
\]
Joint-ranking negatives are constructed by separately perturbing the relation or object while keeping the subject and timestamp fixed. The generation checkpoint is reused for entity activation, whereas rare knowledge forecasting and knowledge obsolescence recognition use original-objective checkpoints, with separate interval fine-tuning where applicable. Rule-based and training-free methods retain their native procedures.

We adapt embedding-based and rule-based baselines into the unified timestamp-level fact-scoring interface
$f_{\theta}(s,r,o,t\mid\mathcal{E}_{t-})$.
Embedding-based baselines directly provide the score of the candidate fact. For rule-based baselines, we use the average confidence of matched temporal rules and assign zero when no rule matches, enabling evaluation through the same ranking interface.

For LLM-based baselines, we follow their official implementations whenever applicable. In rare knowledge forecasting and knowledge obsolescence recognition, LLMs re-rank candidate entities produced by the corresponding retrievers. In knowledge generation, we replace the completion query $(s,r,?,t)$ with $(s,?,?,t)$ while retaining the remaining prompt template. Retrieval is also relation-unconditioned: neither the retriever nor the prompt receives the ground-truth future relation.

For entity activation estimation, LLM-based baselines retain their original candidate-generation and scoring pipelines. Their final re-ranking scores, as defined by the original methods, are aggregated using the same fact-to-source protocol. To reduce direct entity-name cues, we construct prompts using the conceptual entity descriptions (e.g., ``Argentine football player'') rather than exact names.

We implement knowledge obsolescence recognition as a ranking task to retain compatibility with embedding-based and rule-based models, which natively provide timestamp fact scores rather than interval validity decisions. Deriving expiration timestamps from heterogeneous score sequences would require additional model-specific thresholds or change-point heuristics. Ranking valid and obsolete objects within the same candidate space avoids introducing these separate decisions.

\begin{figure}[t]
\centering
\includegraphics[width=0.9\linewidth]{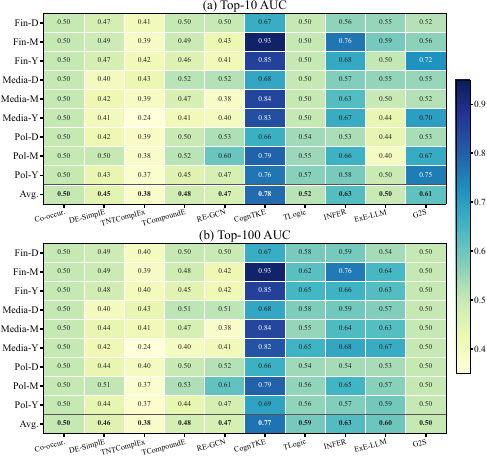}
\caption{Entity activation estimation results on all datasets. Each cell reports ROC-AUC for distinguishing active entities from inactive entities. Top-10 and Top-100 denote activation scores aggregated from the highest-scoring 10 and 100 generated facts, respectively. }
\label{fig:app-activation-full}
\end{figure}

\textbf{Entity Activation Estimation.}
Entity activation reuses the knowledge-generation checkpoint and scoring pipeline. No activation-specific training is performed. At each test timestamp, we select up to $500$ historically frequent subjects that have no observed interaction at the current timestamp or within a granularity-specific look-back window. The window is $180$ steps for day-level datasets, $6$ steps for month-level datasets, and $2$ steps for year-level datasets. These subjects form the inactive set $\mathcal{N}_t$.

Following the activation-score definition in the main paper, we use the highest-scoring ten relation--object pairs, or all available pairs when fewer than ten are returned. For ExE-LLM and G2S, we use their original final re-ranking scores in this aggregation. ROC-AUC is computed within each test timestamp and then averaged across timestamps with nonempty active and inactive sets.

\textbf{Knowledge Generation.}
At each test timestamp, we choose up to $500$ active subjects with the largest numbers of observed test facts. This selection defines the evaluated query population. Future relations and target facts are not supplied to the models.

Given each query $(s,?,?,t)$, scoring-based models rank relation--object pairs from the train-visible vocabularies and retain up to $10{,}000$ predictions. LLM-based methods use their relation-unconditioned retrieval and generation pipelines. Recall is reported at $K\in\{100,1000,5000,10000\}$ and macro-averaged over evaluated positive subject--timestamp queries.

\textbf{Knowledge Obsolescence Recognition.}
For each query $(s,r,?)$, we construct $\mathcal{O}^{\mathrm{obs}}_t(s,r)$ from matching interval facts whose end time is earlier than $t$. Objects in $\mathcal{O}^{+}_t(s,r)$ that remain valid at $t$ are removed from the obsolete set. We evaluate queries with nonempty valid and obsolete object sets. All objects are ranked using the standard fact scorer, without a separate obsolescence-specific scoring head. Valid and obsolete objects use the same raw ranking. Valid-Hits@$N$/Obs-Hits@$N$ and Valid-MRR/Obs-MRR are computed and interpreted jointly following the definitions above.

\textbf{Rare Knowledge Forecasting.}
We set the historical preference size to $K=10$. Models are evaluated using time-filtered ranking. Filtering considers other observed true objects at the same timestamp in the complete test set, rather than only the rare subset.

\textbf{Interval-enhanced Variants.}
In the interval knowledge effectiveness and knowledge obsolescence
recognition experiments, interval facts are expanded within the
training-time range and converted into timestamp facts for
augmentation of the training graph. Trainable models are fine-tuned from the corresponding task-aligned timestamp-only checkpoints: generation checkpoints for the context-effectiveness experiments and original-objective checkpoints for obsolescence recognition. Fine-tuning retains the corresponding task objective. These interval-enhanced variants are reported separately from their timestamp-only counterparts. For history-context-aware models such as RE-GCN, CognTKE, ExE-LLM,
and G2S, expanded interval facts also serve as auxiliary validation
and test history. At query timestamp $t$, only expanded facts
preceding $t$ are used as context. Interval facts are excluded from
timestamp-query construction. The queried interval fact and its equivalent inverse representations are held out from auxiliary context. Historical preference sets and shortcut-controlled subsets are also computed from the original event stream, excluding expanded interval facts. Note that timestamp expansion serves as a compatibility adapter for existing timestamp-based models, not as an interval-native representation. Consequently, these experiments assess the utility of interval-derived auxiliary facts under this adapter rather than models’ intrinsic ability to reason over intervals.

\section{More Experimental Results}
\label{app:more_exp}
\textbf{Entity Activation Estimation}.
Figure~\ref{fig:app-activation-full} compares activation ROC-AUC when
the source score is aggregated from the top-10 and top-100 generated
facts. The two heatmaps show that increasing $K$ does not uniformly
improve all models. CognTKE remains consistently strong under both
settings. INFER and ExE-LLM often improve from Top-10 to Top-100,
indicating that their activation discrimination can benefit from
aggregating scores over a wider candidate set. In contrast, embedding-based models such as DE and TNT show limited activation discrimination under both aggregation settings. G2S can achieve high Top-10 AUC on coarse-grained datasets, but its performance drops toward chance under Top-100 aggregation. This shows that the usefulness of its activation scores is sensitive to the aggregation budget. Overall, the Top-$K$ comparison reveals differences in how candidate scores support active--inactive discrimination.

\begin{table*}[t]
\centering
\small
\setlength{\tabcolsep}{0.8mm}
\begin{tabular}{@{}lccccccccc@{}}
\toprule
Dataset & Co-occur. & DE & TNT & RE-GCN & CognTKE & TLogic & INFER & ExE-LLM & G2S \\
\midrule
Fin.-D & 0.643/0.738 & 0.398/0.683 & 0.104/0.316 & 0.571/0.671 & 0.626/0.736 & 0.317/0.493 & \best{0.898/0.924} & 0.584/0.709 & \second{0.874/0.942} \\
Fin.-M & 0.490/0.621 & 0.184/0.491 & 0.076/0.150 & 0.552/0.682 & 0.513/0.658 & 0.275/0.441 & \second{0.783/0.830} & 0.373/0.498 & \best{0.793/0.873} \\
Fin.-Y & 0.418/0.543 & 0.067/0.282 & 0.013/0.019 & 0.264/0.355 & 0.338/0.421 & 0.227/0.380 & \second{0.602/0.668} & 0.168/0.288 & \best{0.678/0.781} \\
\midrule
Med.-D & 0.620/0.719 & 0.250/0.577 & 0.044/0.143 & 0.571/0.672 & 0.583/0.707 & 0.343/0.532 & \best{0.869/0.916} & 0.482/0.650 & \second{0.809/0.858} \\
Med.-M & 0.513/0.613 & 0.058/0.318 & 0.029/0.114 & 0.400/0.547 & 0.507/0.672 & 0.261/0.488 & \second{0.779/0.823} & 0.288/0.430 & \best{0.789/0.857} \\
Med.-Y & 0.385/0.509 & 0.061/0.287 & 0.030/0.045 & 0.162/0.228 & 0.268/0.369 & 0.218/0.380 & \second{0.619/0.677} & 0.156/0.300 & \best{0.657/0.758} \\
\midrule
Pol.-D & 0.603/0.731 & 0.308/0.641 & 0.040/0.120 & 0.529/0.621 & 0.595/0.651 & 0.384/0.580 & \best{0.857/0.921} & 0.477/0.638 & \second{0.814/0.853} \\
Pol.-M & 0.502/0.608 & 0.176/0.477 & 0.003/0.035 & 0.421/0.556 & 0.503/0.601 & 0.327/0.504 & \best{0.747/0.779} & 0.288/0.425 & \second{0.634/0.726} \\
Pol.-Y & 0.423/0.526 & 0.155/0.455 & 0.002/0.009 & 0.382/0.491 & 0.320/0.378 & 0.303/0.510 & \second{0.568/0.651} & 0.151/0.216 & \best{0.684/0.781} \\
\midrule
Avg. & 0.511/0.623 & 0.184/0.468 & 0.038/0.106 & 0.428/0.536 & 0.473/0.577 & 0.295/0.479 & \second{0.747/0.799} & 0.330/0.462 & \best{0.748/0.825} \\
\bottomrule
\end{tabular}
\caption{Knowledge generation results on all nine datasets. Each cell reports Recall@5000 and Recall@10000. Best and second-best results are marked in bold and underline, ranked by Recall@5000.}
\label{tab:app-generation-transductive-full}
\end{table*}

\textbf{Knowledge Generation}.
Table~\ref{tab:app-generation-transductive-full} reports Recall@5000
and Recall@10000 on the knowledge generation task. Although larger
cutoffs increase the number of returned predictions, most models
remain below the simple co-occurrence baseline on average. CognTKE
recovers more future facts at these larger budgets,
suggesting that some correct targets occur deeper in its ranked
output. Such coverage does not necessarily translate into strong
performance at smaller generation budgets. G2S and INFER achieve
the highest average Recall@5000, with closely matched scores of
0.748 and 0.747, respectively. These results distinguish broad
target coverage from accurate top-ranked retrieval.

\begin{table*}[t]
\centering
\small
\setlength{\tabcolsep}{0.8mm}
\begin{tabular}{@{}lccccccccc@{}}
\toprule
Dataset & Co-occur. & DE & TNT & RE-GCN & CognTKE & TLogic & INFER & ExE-LLM & G2S \\
\midrule
Fin.-D & 0.072/0.011 & 0.266/0.320 & 0.308/0.392 & 0.070/0.008 & 0.051/0.006 & 0.049/0.000 & 0.324/0.306 & 0.081/0.047 & 0.058/0.006 \\
Fin.-M & 0.112/0.014 & 0.329/0.314 & 0.331/0.277 & 0.045/0.045 & 0.095/0.006 & 0.127/0.041 & 0.385/0.255 & 0.127/0.039 & 0.094/0.008 \\
Fin.-Y & 0.060/0.012 & 0.232/0.254 & 0.229/0.223 & 0.007/0.002 & 0.120/0.013 & 0.130/0.049 & 0.316/0.324 & 0.107/0.044 & 0.037/0.003 \\
\midrule
Med.-D & 0.079/0.012 & 0.310/0.354 & 0.324/0.381 & 0.062/0.008 & 0.095/0.005 & 0.049/0.000 & 0.309/0.309 & 0.081/0.059 & 0.067/0.018 \\
Med.-M & 0.107/0.009 & 0.336/0.361 & 0.376/0.367 & 0.072/0.008 & 0.093/0.007 & 0.119/0.040 & 0.371/0.238 & 0.104/0.025 & 0.091/0.011 \\
Med.-Y & 0.062/0.011 & 0.257/0.272 & 0.246/0.256 & 0.027/0.002 & 0.147/0.041 & 0.138/0.054 & 0.297/0.300 & 0.093/0.044 & 0.059/0.006 \\
\midrule
Pol.-D & 0.023/0.015 & 0.475/0.585 & 0.634/0.726 & 0.006/0.000 & 0.030/0.004 & 0.000/0.000 & 0.422/0.342 & 0.047/0.015 & 0.023/0.012 \\
Pol.-M & 0.073/0.024 & 0.256/0.373 & 0.402/0.634 & 0.001/0.000 & 0.061/0.007 & 0.138/0.067 & 0.322/0.360 & 0.068/0.015 & 0.085/0.040 \\
Pol.-Y & 0.054/0.019 & 0.096/0.070 & 0.323/0.420 & 0.001/0.000 & 0.148/0.063 & 0.113/0.068 & 0.269/0.308 & 0.042/0.018 & 0.053/0.018 \\
\midrule
Avg. & 0.071/0.014 & 0.284/0.323 & 0.352/0.408 & 0.032/0.008 & 0.093/0.017 & 0.096/0.035 & 0.335/0.305 & 0.083/0.034 & 0.063/0.014 \\
\bottomrule
\end{tabular}
\caption{Knowledge obsolescence recognition results on all TKGB datasets. Each cell reports Valid-Hits@10 and Obs-Hits@10. Valid-Hits@10 measures whether still-valid interval objects are retained in the top-10 predictions, while Obs-Hits@10 measures whether obsolete objects are incorrectly ranked in the top-10. Higher Valid-Hits@10 and lower Obs-Hits@10 indicate better obsolescence-aware forecasting. Query interval facts are held out from the temporal context.}
\label{tab:obsolescence-suppression-appendix-full}
\end{table*}

\begin{table*}[t]
\centering
\small
\setlength{\tabcolsep}{0.8mm}
\begin{tabular}{@{}lccccccccc@{}}
\toprule
Dataset & Co-occur. & DE & TNT & RE-GCN & CognTKE & TLogic & INFER & ExE-LLM & G2S \\
\midrule
Fin.-D & 0.040/0.007 & 0.128/0.173 & {0.150/0.179} & 0.039/0.007 & 0.037/0.002 & 0.027/0.001 & {0.215/0.215} & 0.048/0.023 & 0.036/0.005 \\
Fin.-M & 0.059/0.006 & 0.143/0.161 & {0.163/0.140} & 0.023/0.020 & 0.047/0.005 & 0.069/0.012 & {0.237/0.152} & 0.079/0.021 & 0.062/0.005 \\
Fin.-Y & 0.037/0.004 & {0.100/0.122} & 0.099/0.096 & 0.004/0.001 & 0.059/0.008 & 0.057/0.021 & {0.170/0.194} & 0.058/0.017 & 0.025/0.002 \\
\midrule
Med.-D & 0.038/0.007 & 0.149/0.194 & {0.170/0.185} & 0.028/0.006 & 0.051/0.002 & 0.027/0.001 & {0.202/0.213} & 0.046/0.025 & 0.033/0.011 \\
Med.-M & 0.051/0.006 & 0.152/0.190 & {0.181/0.173} & 0.033/0.004 & 0.055/0.005 & 0.056/0.019 & {0.214/0.140} & 0.077/0.021 & 0.048/0.005 \\
Med.-Y & 0.039/0.004 & 0.109/0.130 & {0.114/0.107} & 0.012/0.001 & 0.079/0.016 & 0.057/0.022 & {0.165/0.165} & 0.050/0.017 & 0.035/0.003 \\
\midrule
Pol.-D & 0.010/0.005 & {0.285/0.308} & {0.356/0.391} & 0.004/0.002 & 0.028/0.001 & 0.002/0.001 & 0.281/0.261 & 0.026/0.013 & 0.011/0.005 \\
Pol.-M & 0.041/0.007 & 0.107/0.191 & {0.184/0.367} & 0.001/0.001 & 0.035/0.005 & 0.057/0.040 & {0.172/0.250} & 0.041/0.011 & 0.042/0.010 \\
Pol.-Y & 0.029/0.006 & 0.044/0.042 & {0.161/0.227} & 0.002/0.001 & 0.074/0.033 & 0.048/0.036 & {0.136/0.192} & 0.027/0.010 & 0.028/0.005 \\
\midrule
Avg. & 0.038/0.006 & 0.135/0.168 & {0.175/0.207} & 0.016/0.005 & 0.052/0.009 & 0.044/0.017 & {0.199/0.198} & 0.050/0.018 & 0.036/0.006 \\
\bottomrule
\end{tabular}
\caption{Knowledge obsolescence recognition results on all nine datasets. Each cell reports Valid-MRR/Obs-MRR. Valid-MRR measures whether still-valid interval objects are ranked high, while Obs-MRR measures whether obsolete objects are incorrectly ranked high. Higher Valid-MRR and lower Obs-MRR are preferred.}
\label{tab:ranking-obsolescence-mrr}
\end{table*}

\textbf{Knowledge Obsolescence Recognition}.
Tables~\ref{tab:obsolescence-suppression-appendix-full} and
\ref{tab:ranking-obsolescence-mrr} provide the paired valid and
obsolete results over all nine datasets. Models that retrieve more
valid interval objects often also rank more obsolete objects highly.
For example, TNT achieves average Valid-Hits@10/Obs-Hits@10 of
0.352/0.408. Conversely, RE-GCN obtains 0.032/0.008, indicating that
both valid and obsolete objects are rarely retrieved. Its low
Obs-Hits@10 therefore cannot, by itself, demonstrate effective
obsolescence recognition. The same joint interpretation applies
to Valid-MRR/Obs-MRR. Results also vary across temporal granularities,
showing that retaining valid knowledge while reducing obsolete-object
ranking remains challenging.

\begin{table*}[t]
\centering
\small
\setlength{\tabcolsep}{1mm}
\begin{tabular}{@{}lcccccccc@{}}
\toprule
Dataset & DE & TNT & RE-GCN & CognTKE & TLogic & INFER & ExE-LLM & G2S \\
\midrule
Fin.-D & 0.001/0.003 & 0.000/0.002 & 0.003/0.010 & \best{0.330/0.464} & 0.035/0.056 & \second{0.074/0.092} & 0.000/0.037 & 0.008/0.016 \\
Fin.-M & 0.001/0.005 & 0.001/0.006 & 0.005/0.019 & \second{0.016/0.040} & 0.007/0.027 & \best{0.048/0.059} & 0.011/0.032 & 0.014/0.032 \\
Fin.-Y & 0.008/0.027 & 0.007/0.025 & 0.014/0.027 & 0.010/0.027 & \best{0.019/0.052} & \second{0.015/0.030} & 0.003/0.022 & 0.006/0.025 \\
\midrule
Med.-D & 0.001/0.005 & 0.001/0.003 & 0.004/0.015 & \best{0.336/0.475} & 0.027/0.052 & \second{0.069/0.085} & 0.009/0.026 & 0.004/0.013 \\
Med.-M & 0.001/0.005 & 0.001/0.005 & 0.006/0.021 & 0.009/0.048 & 0.009/0.034 & \best{0.045/0.055} & 0.003/0.023 & \second{0.010/0.027} \\
Med.-Y & 0.008/0.024 & 0.007/0.023 & 0.010/0.032 & 0.007/0.023 & \best{0.015/0.048} & \second{0.015/0.031} & 0.006/0.027 & 0.012/0.024 \\
\midrule
Pol.-D & 0.002/0.009 & 0.001/0.007 & 0.005/0.022 & \best{0.322/0.453} & 0.014/0.037 & \second{0.028/0.061} & 0.004/0.017 & 0.012/0.033 \\
Pol.-M & 0.001/0.007 & 0.001/0.008 & 0.006/0.022 & \best{0.049/0.080} & 0.009/0.031 & \second{0.022/0.044} & 0.000/0.007 & 0.000/0.014 \\
Pol.-Y & 0.005/0.018 & 0.005/0.018 & 0.007/0.016 & 0.004/0.015 & \best{0.015/0.040} & 0.009/0.024 & 0.006/0.025 & \second{0.009/0.009} \\
\midrule
Avg. & 0.003/0.012 & 0.003/0.011 & 0.007/0.020 & \best{0.120/0.181} & 0.017/0.042 & \second{0.036/0.054} & 0.005/0.024 & 0.009/0.022 \\
\bottomrule
\end{tabular}
\caption{Rare knowledge forecasting results. Each cell reports Hits@1 and Hits@3, and the best results are highlighted according to Hits@1.}
\label{tab:app-exp1-hard-full}
\end{table*}

\textbf{Rare Knowledge Forecasting}.
Table~\ref{tab:app-exp1-hard-full} reports extended rare knowledge
forecasting results. The results are consistent with
our previous analysis and further reveal substantial variation
across temporal granularities. In particular, CognTKE achieves
rare-query Hits@1 of 0.322--0.336 on day-level datasets, compared
with 0.009--0.049 on month-level datasets and 0.004--0.010 on
year-level datasets. Strong day-level performance therefore does
not imply uniformly strong rare-query forecasting at coarser
resolutions. Reporting these slices separately exposes differences
that an overall average can obscure.

\begin{table*}[t]
\centering
\small
\setlength{\tabcolsep}{0.8mm}
\begin{tabular}{@{}lccccccccc@{}}
\toprule
Dataset & Co-occur. & DE & TNT & RE-GCN & CognTKE & TLogic & INFER & ExE-LLM & G2S \\
\midrule
Fin.-D & 0.500/0.500 & 0.453/0.481 & 0.388/0.398 & 0.437/0.424 & \best{0.891/0.891} & 0.500/0.526 & 0.356/0.357 & \second{0.629/0.629} & 0.609/0.605 \\
Fin.-M & 0.500/0.500 & 0.424/0.429 & 0.399/0.394 & 0.376/0.375 & 0.394/0.393 & 0.500/0.560 & 0.381/0.524 & \best{0.669/0.669} & \second{0.615/0.621} \\
\midrule
Med.-D & 0.500/0.500 & 0.415/0.415 & 0.393/0.399 & \second{0.573/0.548} & \best{0.924/0.924} & 0.500/0.528 & 0.394/0.364 & 0.555/0.555 & 0.533/0.529 \\
Med.-M & 0.500/0.500 & 0.470/0.482 & 0.381/0.385 & 0.457/0.466 & 0.454/0.448 & 0.500/0.597 & 0.425/0.542 & \best{0.612/0.612} & \second{0.549/0.551} \\
\midrule
Pol.-D & 0.500/0.500 & 0.483/0.482 & 0.417/0.410 & \second{0.550/0.526} & \best{0.790/0.788} & 0.500/0.532 & 0.474/0.476 & 0.513/0.513 & 0.507/0.507 \\
Pol.-M & 0.500/0.500 & 0.447/0.446 & 0.384/0.384 & 0.471/0.479 & \second{0.572/0.571} & 0.500/0.551 & 0.397/0.432 & \best{0.624/0.624} & 0.564/0.565 \\
\midrule
Avg. & 0.500/0.500 & 0.449/0.456 & 0.394/0.395 & 0.477/0.470 & \best{0.671/0.669} & 0.500/0.549 & 0.405/0.449 & \second{0.600/0.600} & 0.563/0.563 \\
\bottomrule
\end{tabular}
\caption{Online entity activation results. Each cell reports AUC for distinguishing active entities using the activation scores aggregated from the highest-scoring 10 and 100 generated facts. Best and second-best results are marked in bold and underline.}
\label{tab:online-activation}
\end{table*}

\textbf{Online Testing Performance}.
Table~\ref{tab:online-activation} reveals that strong generation
performance does not necessarily imply strong source-activity
discrimination: G2S has lower average activation AUC than CognTKE
and ExE-LLM. CognTKE achieves the highest average AUC under both
aggregation budgets. Co-occur. remains at 0.500, while TLogic
improves from 0.500 under Top-10 aggregation to 0.549 under Top-100.
Together, these results show that generation coverage and activation
discrimination capture complementary aspects of forecasting
performance on newly collected events.

\end{document}